\documentclass[]{spie}

\usepackage{float}
\usepackage{tabularx}
\usepackage{graphicx,graphics}
\usepackage{url}
\usepackage{amsmath,amsfonts,amssymb,textcomp,booktabs,threeparttable}
\usepackage{amssymb}
\usepackage{hhline}
\usepackage{multirow}
\usepackage[bookmarks=true,bookmarksopen=true,breaklinks=true,colorlinks=true, linkcolor=blue,anchorcolor=blue,citecolor=blue,filecolor=blue,menucolor=blue, urlcolor=blue,bookmarks=false]{hyperref}
\usepackage{stmaryrd}
\include{boldcommands}
\usepackage{tabularx}
\usepackage{booktabs}
\usepackage{subcaption}
\usepackage{eucal,epstopdf}

\title{Distributed deep learning for robust multi-site segmentation of CT imaging after traumatic brain injury}

\author[1,2,3,4]{Samuel Remedios}
\author[1,2]{Snehashis Roy}
\author[4]{Justin Blaber}
\author[5]{Camilo Bermudez}
\author[6]{Vishwesh Nath}
\author[7]{Mayur B. Patel}
\author[2]{John A. Butman}
\author[4,5,6]{Bennett A. Landman}
\author[1,2]{Dzung L. Pham}

\affil[1]{Center for Neuroscience and Regenerative Medicine, Henry Jackson Foundation}
\affil[2]{Radiology and Imaging Sciences, Clinical Center, National Institute of Health}
\affil[3]{Department of Computer Science, Middle Tennessee State University}
\affil[4]{Department of Electrical Engineering, Vanderbilt University}
\affil[5]{Department of Biomedical Engineering, Vanderbilt University}
\affil[6]{Department of Computer Science, Vanderbilt University}
\affil[7]{Departments of Surgery, Neurosurgery, Hearing \& Speech Sciences; Center for Health Services Research, Vanderbilt Brain Institute; Critical Illness, Brain Dysfunction, and Survivorship Center, Vanderbilt University Medical Center; VA Tennessee Valley Healthcare System, Department of Veterans Affairs Medical Center}

\begin{document}
\maketitle

\begin{abstract}
Machine learning models are becoming commonplace in the domain of medical imaging, and with these methods comes an ever-increasing need for more data.  However, to preserve patient anonymity it is frequently impractical or prohibited to transfer protected health information (PHI) between institutions.  Additionally, due to the nature of some studies, there may not be a large public dataset available on which to train models.  To address this conundrum, we analyze the efficacy of transferring the model itself in lieu of data between different sites.  By doing so we accomplish two goals: $1$) the model gains access to training on a larger dataset that it could not normally obtain and $2$) the model better generalizes, having trained on data from separate locations.  In this paper, we implement multi-site learning with disparate datasets from the National Institutes of Health (NIH) and Vanderbilt University Medical Center (VUMC) without compromising PHI.  Three neural networks are trained to convergence on a computed tomography (CT) brain hematoma segmentation task: one only with NIH data, one only with VUMC data, and one multi-site model alternating between NIH and VUMC data.  Resultant lesion masks with the multi-site model attain an average Dice similarity coefficient of $0.64$ and the automatically segmented hematoma volumes correlate to those done manually with a Pearson correlation coefficient of $0.87$, corresponding to an $8$\% and $5$\% improvement, respectively, over the single-site model counterparts.

\keywords{multi-site, distributed, deep learning, neural network, computed tomography (CT), hematoma, lesion, segmentation}
\end{abstract}

\section{Introduction}
\label{sec:intro}
Deep learning has recently become a key approach for computer vision and medical imaging problems.  Neural networks have been used to skull-strip CT scans\cite{akkus2018extraction}, segment magnetic resonance images\cite{pereira2016brain}, locate and segment blood vessels\cite{liskowski2016segmenting}, as well as segment brain regions\cite{de2015deep} and lesions\cite{kamnitsas2015multi}.  A wide variety of models and architectures have been implemented to solve these tasks, and there also exist pre-trained models prepared for general use cases\cite{lin2013network}.  Regardless of the particular task for which a model is designed or selected, machine learning methods generally benefit from the inclusion of more data for training and validating the model\cite{halevy2009unreasonable}.  Traditionally, acquisition of multi-site data involves data transfer to a centralized location on which the desired model trains; however, it is frequently prohibited or difficult to acquire HIPAA-compliant health data transfer permits\cite{nihDataShare}.  These data restrictions are vital, though, as protected health information (PHI) policies enforce respect for patient privacy and anonymity.\cite{luxton2012mhealth}$^,$\cite{thompson2011protected}$^,$\cite{fetzer2008hipaa}  Herein lies a contradiction: machine learning models benefit greatly from a wealth of data, yet datasets related to healthcare cannot be shared between sites easily.

To address this problem, we propose to transfer the models themselves between sites in lieu of a dataset transfer.  The concept of distributed learning is not new to machine learning, with one such example coming from Google's implementation of Federated Learning, through which models are averaged between mobile phones\cite{konevcny2016federated}. However, this approach does not have the goal of gaining accuracy or generalizability, and instead is a decentralized framework geared towards mobile devices and their limited computing power.  Another distributed learning technique is transfer learning\cite{pan2010survey}, which aims to apply useful features learned from one task towards a kick-started learning for some other task.  Different still is the concept of asynchronous stochastic gradient descent\cite{zhang2013asynchronous}, wherein a model is copied for some number of splits of training data, and their learned weights are aggregated once training is complete.

Recently, a study has embarked to investigate whether a model can perform better if it accesses data from different sites \cite{chang2018distributed}, wherein the authors simulate a multi-site scenario by splitting an open-source dataset into groups and apply different transformations and noise to each group with the goal of making the data appear different.  The authors investigate applying different multi-site training approaches, comparing transfer learning to different patterns of passing partially trained models.

\begin{figure*}[t] 
\begin{center}
\includegraphics[width=1\textwidth]{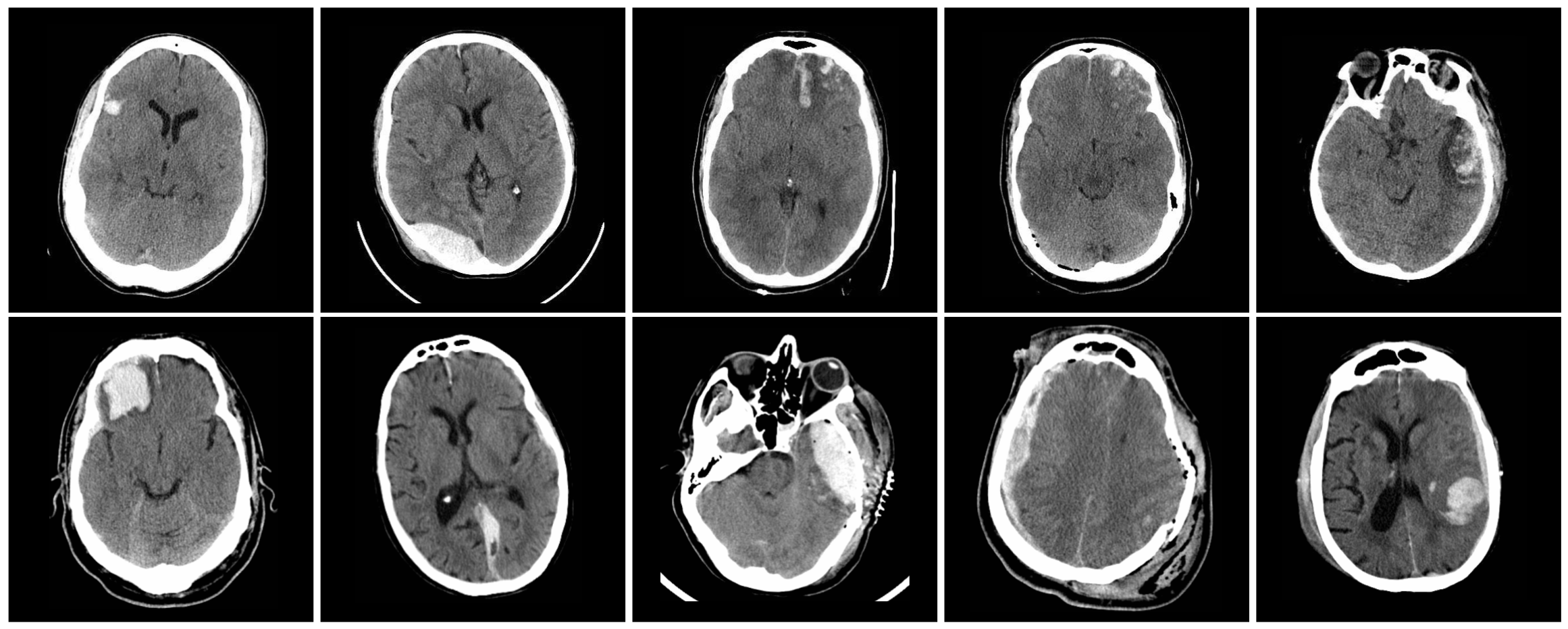}
\end{center}
\caption{Representative $5.0$ mm thick transverse CT sections through the head in $10$ subjects with TBI.  In-plane resolution is approximately $0.5\times0.5$ mm.  In each case, the hemorrhagic lesion appears intermediate density between normal brain tissue and bone. Note the heterogeneity of size, location, density and configuration.}
\label{fig:Fig1}
\end{figure*}

In this paper, we expand upon this work by using empirical multi-site data, separately acquired from the NIH and VUMC.  Because of differences in the acquisition at each site, as well as in delineation protocols, improved performance due to the combined training data gained by multi-site learning is not guaranteed.  Thus, we employ the aforementioned paper's cyclic weight transfer as our training paradigm and forgo the uni-directional transfer learning approach.

Our specific contributions are the presentation of an extensible framework through which multiple sites can train the same model using private data and the validation of the efficacy of two different training schema on the segmentation of hematoma in traumatic brain injury (TBI) CT scans.  In the latter contribution, we consider single-site learning at each of the two sites (NIH and VUMC), and multi-site learning between both sites.

Here, we target segmentation of hemorrhages and hematomas in patients with TBI (see Figure~\ref{fig:Fig1}).  Hemorrhages refer to active bleeding, while a hematoma is any collection or swelling of clotted blood outside of the blood vessels, the cause of which could be severe trauma or disease.  The identification and segmentation of blood is an important consideration for diagnosis, prediction of patient recovery, and for examining correlations with long-term neurologic disabilities\cite{trifan2017mr} such as cognitive impairment\cite{kinnunen2010white}. Improving the efficacy of hematoma segmentation will therefore assist developments in understanding and treating TBI.

\begin{table}[tb]
\caption{Distribution of CT image volumes between training and test sets for both sites.
}
\label{tab:data}
\begin{center}
\begin{tabular}{lll}
\toprule[2pt]
Training Location & \# Training Images & \# Testing Images\\
\toprule[1pt]
VUMC & $10$ & $8$\\ 
\cmidrule[1pt]{2-3}
NIH & $17$ & $10$\\
\toprule[1pt]
Total & $\mathbf{27}$ & $\mathbf{18}$\\ 
\bottomrule[2pt]
\end{tabular}
\end{center}
\vspace{-1em}
\end{table}

\section{Method}
\label{sec:method}
\subsection{Data}

CT images from $27$ acute TBI patients presenting with intracranial hematomas were acquired as part of a research study by the Center for Neuroscience and Regenerative Medicine (CNRM) and NIH.  At VUMC, $18$ CT images of TBI patients were obtained in de-identified form.  The resolutions of all scans from both sites were approximately $0.5\times0.5\times5.0$ mm$^3$.  All scans were converted from DICOM to NIFTI and subsequently transformed into Hounsfield units.  For training, $10$ scans were used at the VUMC site while $17$ were used at the NIH; the remaining $8$ and $10$, respectively, were set aside as the test dataset.  Images from both the NIH and VUMC had a variety of hematoma types, sizes, and locations; however VUMC on average had a larger hematoma volume of $41,000$ mm$^3$ compared with $13,700$ mm$^3$ in the NIH dataset. For preprocessing, all CT image volumes underwent skull-stripping by \texttt{CT\_BET} \cite{muschelli2015validated} and were rigidly transformed to a common orientation. To address the low number of training images, we collected $1,000$ $255\times255$ 2D patches from each CT volume, $20\%$ of which were used as a validation set for hyperparameter tuning.  Since voxel intensities were in Hounsfield units, no normalization was applied and thus no intensities were scaled. Additionally, because the images have low through-plane resolution ($5.0$ mm) compared to the in-plane resolution ($0.5$ mm), only 2D segmentations were considered.  Manual segmentations were performed by independent raters at the two sites and reviewed independently by a neuroradiologist; quantities are reflected in Table~\ref{tab:data}. 

\begin{figure}[tbh] 
\begin{center}
\includegraphics[width=0.75\textwidth]{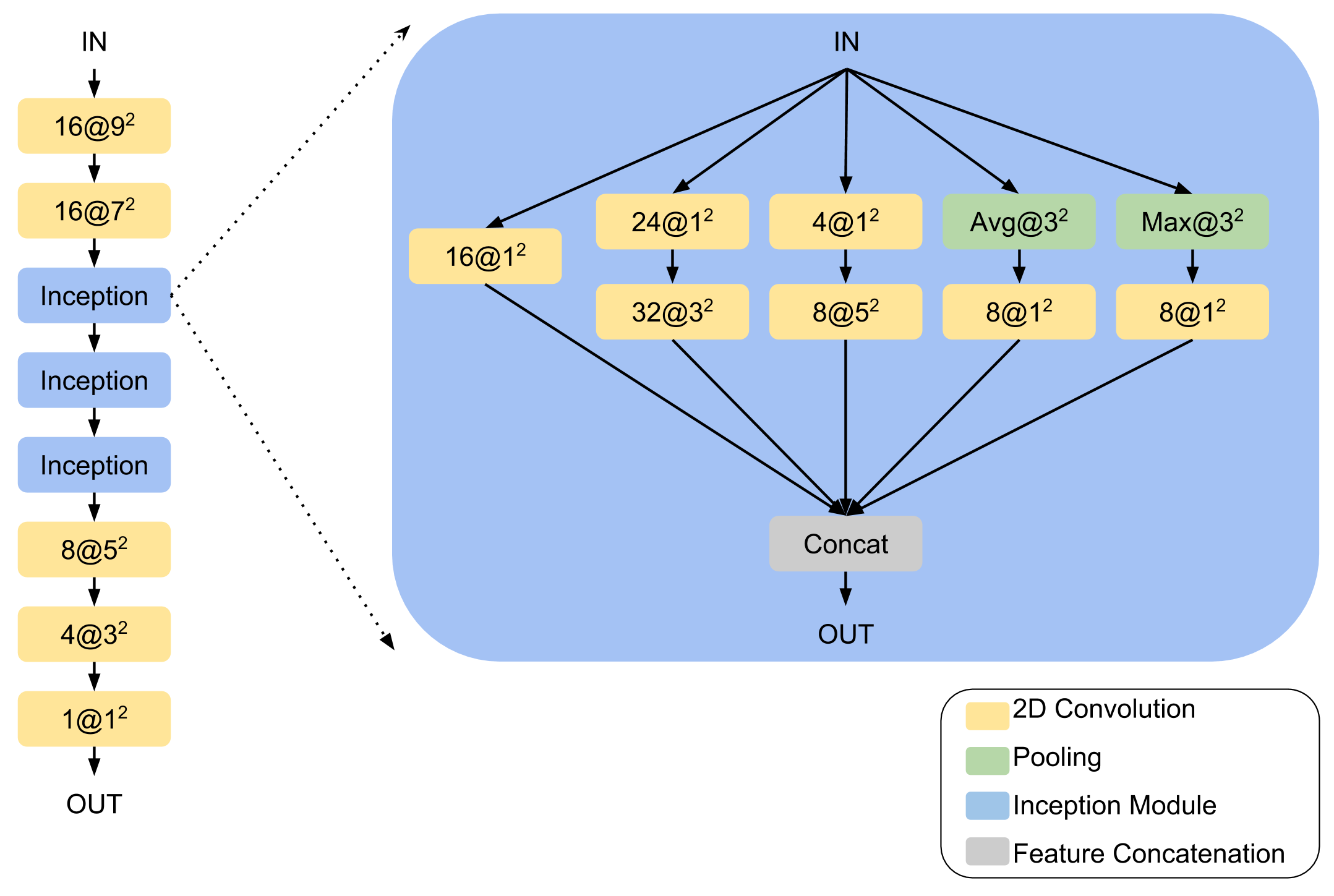}
\end{center}
\vspace{-1em}
\caption{Illustrated is the proposed model architecture.  Convolution layers are indicated in yellow, with notation $N\ @\ k^2$ representing $N$ $2D$ kernels of size $k\times k$.  The activation for all convolution layers is \texttt{ReLU}, except for the final $1\ @\ 1^2$ convolution which uses a sigmoid activation.  \texttt{Avg} $@\ k^2$ and \texttt{Max} $@\ k^2$ respectively correspond to average pooling and max pooling with strides $k\times k$.  The modified Inception Module, shown to the right, is a variation of the original Inception Module presented by Google\cite{szegedy2015going}.}
\label{fig:Fig2}
\end{figure}

\subsection{Model Architecture}
Previously, an Inception Net-based architecture has performed well on hematoma segmentation from magnetic resonance images\cite{roy2018tbi}; as such, we utilize a similar 2D architecture with arbitrary-sized inputs, permitting 2D patch-wise training and full slice automatic segmentation.  This architecture is illustrated in Figure~\ref{fig:Fig2}.  Training continued to convergence, defined as no loss improvement of $1\times10^{-4}$ in $10$ epochs on the validation patch set.  The learning rate was set at $1\times10^{-4}$ with the Adam\cite{kingma2014adam} optimizer and the continuous Dice coefficient\cite{shamir2018continuous}(cDC) as the loss.  Resultant binary segmentation masks were generated by thresholding the probability masks at $0.5$.

\subsection{Framework Implementation}
To implement multi-site learning using cyclic weight transfer, we established a server which both the NIH and VUMC could securely access.  On this server we mounted a single directory where the neural network weights were kept.  Identical Python scripts at both institutions allowed the model to be loaded, trained, and saved via secure shell access to this tertiary server without opening up public connections to either institution's data \cite{multisiteLearningGitHub}. 

Particularly, in our implementation, data at each site is never accessible to investigators outside that institution.

\subsection{Training Strategies}

\textbf{Single-Site Learning} As a baseline, each of the sites NIH and VUMC performed single-site learning (SSL) to convergence with their respective datasets.  Once converged, each of the NIH SSL and VUMC SSL models were evaluated on the NIH test and VUMC test sets.  Concretely, NIH SSL was trained on the NIH training dataset and tested on the NIH and VUMC testing datasets and VUMC SSL was trained on the VUMC train dataset and tested on both the NIH and VUMC testing datasets.

\textbf{Multi-Site Learning} Multi-site learning (MSL) involved training the same model architecture from initialization (i.e.: no transfer learning), then passing the model to the next institution for the subsequent epoch.  Thus, MSL would train for one epoch on the NIH train dataset, then one epoch on VUMC train dataset, then one epoch on NIH train dataset, and so on until convergence.  As with the NIH SSL and VUMC SSL models, the MSL model was evaluated over both the NIH and VUMC testing datasets.

\begin{figure}[tb] 
\begin{center}
\includegraphics[width=1\textwidth]{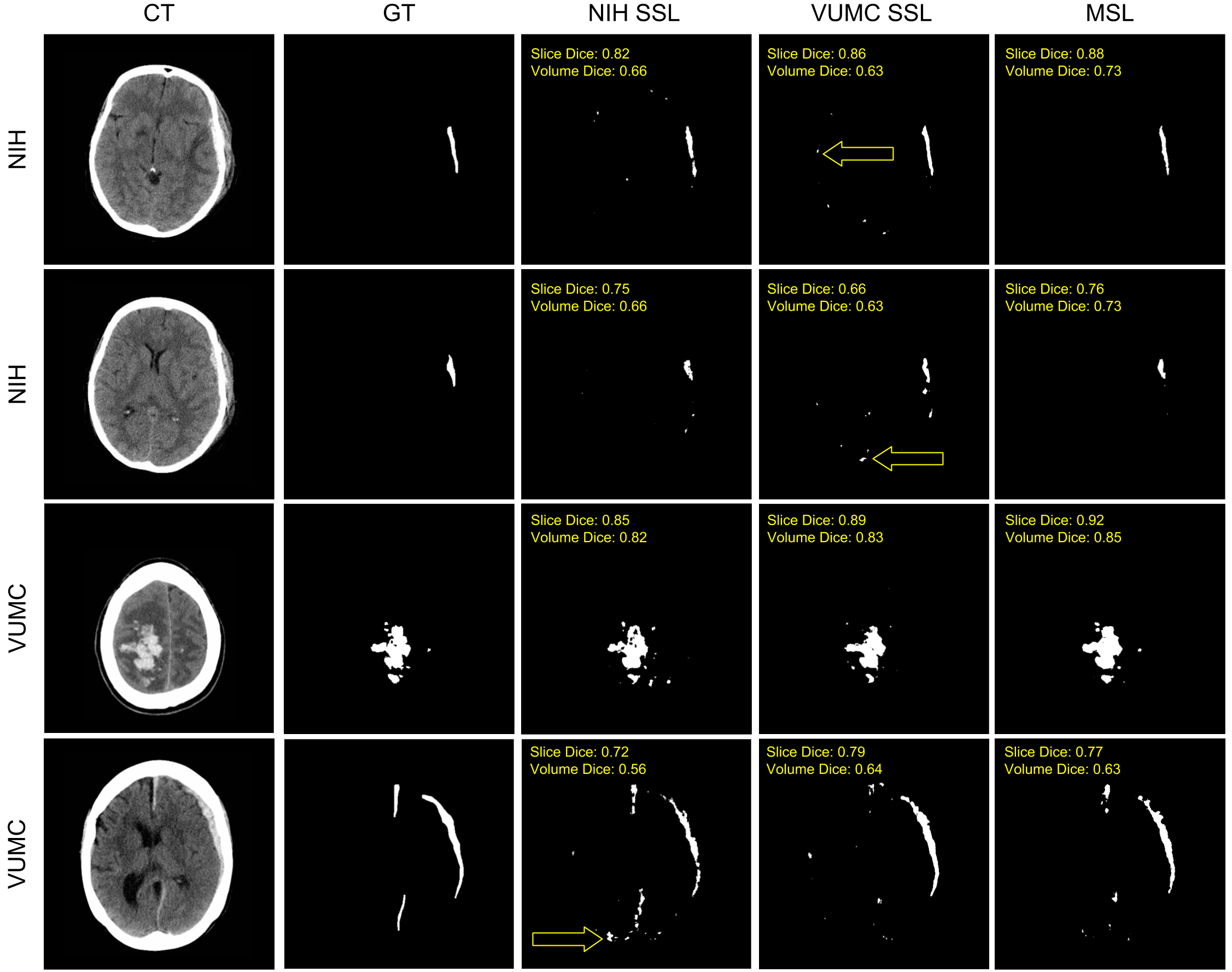}
\end{center}
\caption{Examples of automatic segmentations are shown.  From left to right, the images correspond to the CT, the manual ``ground truth" (GT) segmentation, the NIH SSL, VUMC SSL, and MSL segmentations.  Both the image volume and the specific image slice's Dice coefficient are overlaid on that segmentation.  Yellow arrows specify examples of false positives near the blood-brain barrier which were not present in the MSL segmentations.}
\label{fig:predictions}
\end{figure}

\section{Results}
After training, we have three distinct sets of weights for our model: NIH SSL,  VUMC SSL, and MSL.  Each of these was evaluated over both the NIH and VUMC testing datasets.  We validated all weight sets with two quantitative metrics: the Dice coefficient and hematoma volume correlation between the automatic and manual segmentations.  Further explanation of these measurements follows.

\begin{figure}
\centering
\begin{minipage}[t]{.475\textwidth}
  \centering
  \includegraphics[width=1\linewidth]{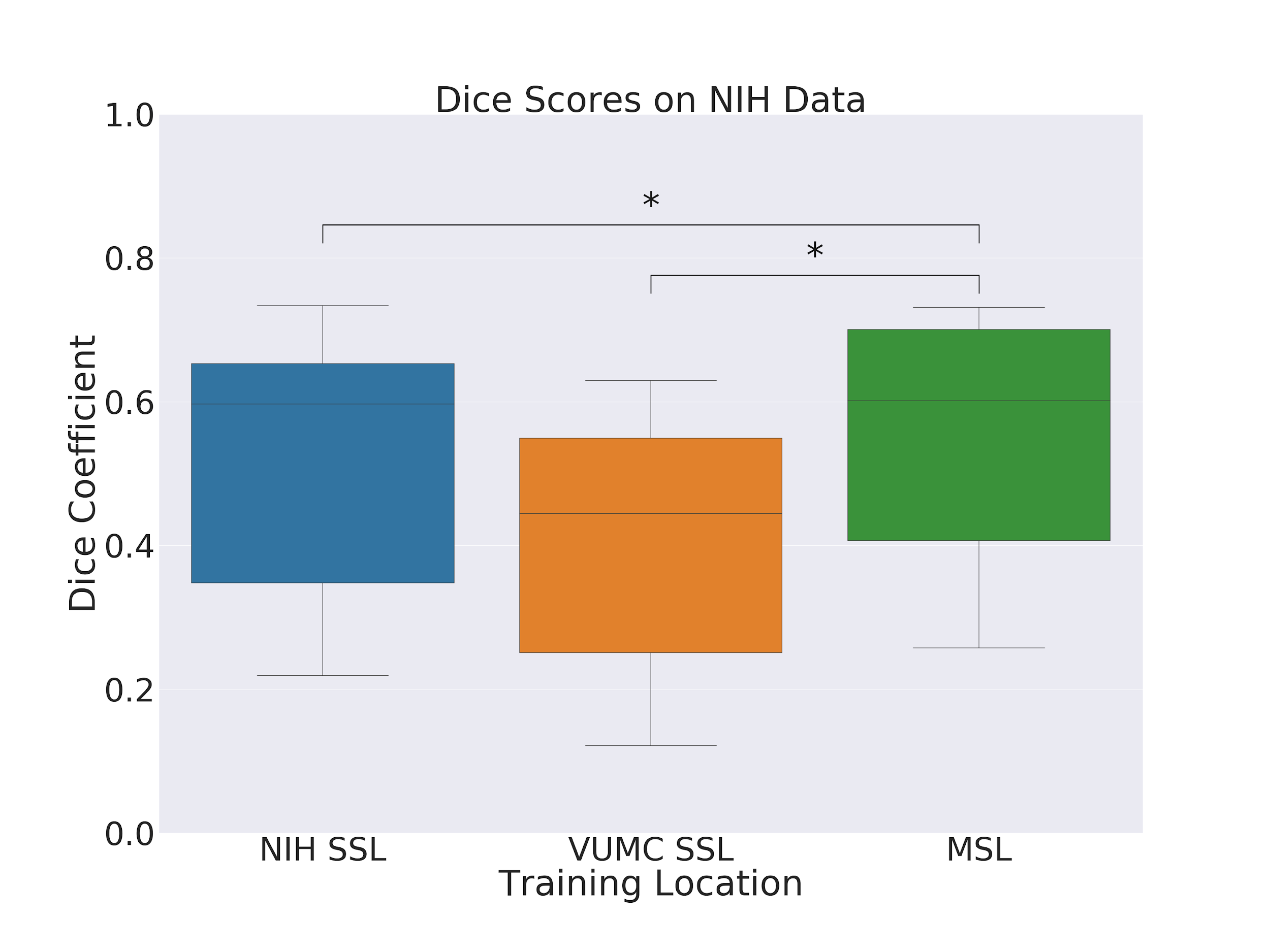}
  \captionof{figure}{Model Dice coefficient comparison over the NIH testing dataset.  The MSL model performed significantly better than either SSL models, where the asterisk indicates a significant difference ($p<0.05$, found via the Wilcoxon signed-rank test).}
  \label{fig:nih_boxplot}
\end{minipage}
\hspace{0.25cm}
\begin{minipage}[t]{0.475\textwidth}
  \centering
  \includegraphics[width=1\linewidth]{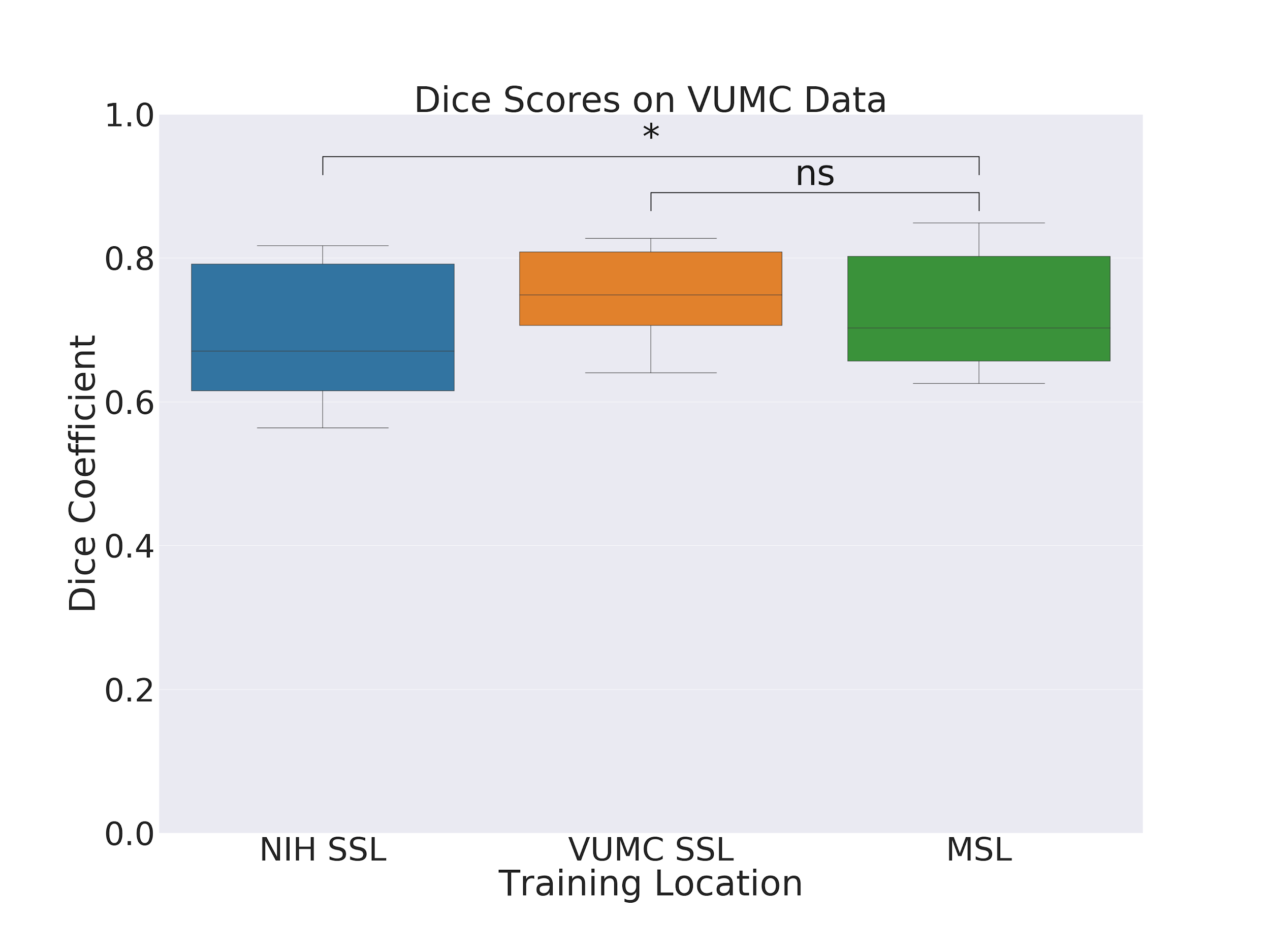}
  \captionof{figure}{Model Dice coefficient comparison over the VUMC testing dataset.  The MSL model performed significantly better than the NIH SSL model, but performed worse than the VUMC SSL model.  The asterisk indicates significance ($p<0.05$, according to the Wilcoxon signed-rank test) where ``ns" corresponds to ``not significant" ($p>=0.05$).}
  \label{fig:vu_boxplot}
\end{minipage}
\end{figure}

\subsection{Qualitative Evaluation}
The automatic segmentations of test CT slices in Fig.~\ref{fig:predictions} allow for qualitative comparisons between the different training sites.  As expected, the model trained at its respective location shows fewer false positives than the model trained at the other location.  However, in these scenarios we see the MSL model generally contains less predicted hematoma voxels.  Yellow arrows indicate false positives not only near the blood-brain barrier, but also ones that are not present in the MSL segmentations.

\subsection{Quantitative Evaluation}
Separate from the cDC loss function, the traditional Dice coefficient was employed to judge the accuracy of the automatic masks.  Figure~\ref{fig:predictions} displays example segmentation results from four different patients while Table~\ref{tab:results} shows the overall averages for all models.

To compare the efficacy of the MSL model against the two SSL models, we used the Wilcoxon signed-rank test over the corresponding Dice scores.  Our findings, illustrated in Figures~\ref{fig:nih_boxplot} and \ref{fig:vu_boxplot}, show significant improvement between the MSL model and both the NIH SSL (p=0.009) and VUMC SSL (p=0.005) models with respect to the NIH test dataset, and a significant improvement between the MSL model and the NIH SSL (p=0.01) over the VUMC test dataset.  The VUMC SSL model outperformed the MSL model on the VUMC test data, but not significantly (p=0.337).

Two considerations are made regarding low Dice scores.  First, specifically regarding the disparity of average Dice scores between the NIH and VUMC visible in Figures~\ref{fig:nih_boxplot} and \ref{fig:vu_boxplot}, data from the NIH had a lower average hematoma volume than VUMC data ($13,700$ mm$^3$ for NIH data versus $41,000$ mm$^3$ for VUMC data), and Dice coefficients between two segmentations are known to be dependent on the volumes of the objects being considered.  Second, regarding overall average Dice scores for both institutions, some 2D image slices near the top and bottom of the brain as well as along the blood-brain barrier suffer from increased false positives.  These are shown in Figure~\ref{fig:predictions}, marked by yellow arrows.

As an alternate means to evaluate the accuracy of the automatic segmentation, we calculated the Pearson correlation coefficient between the total volume in the segmentation and manual masks, as provided in Table~\ref{tab:results}. Although the automatic segmentations contain some small false positives which reduce the Dice coefficient, overall, the volume correlations remain high. 

\begin{table}[tb]
\caption{Average Dice coefficients and Pearson correlation coefficients for the three training strategies over the NIH and VUMC datasets.  The average result over both datasets is shown to illustrate each model's general ability.  An asterisk indicates significant improvements in Dice coefficient ($p < 0.05$) between the MSL and each of the NIH SSL and VUMC SSL models as evaluated by the Wilcoxon signed-rank test, and bold text indicates the highest Pearson correlation coefficient between automatic and manual segmented hematoma volumes.}
\label{tab:results}
\begin{center}
\begin{tabular}{l|cc|cc|cc}
\toprule[2pt]
& \multicolumn{2}{c|}{NIH Data} &  \multicolumn{2}{c|}{VUMC Data} & \multicolumn{2}{c}{Average of NIH and VUMC data} \\
\hline
& Dice & Correlation & Dice & Correlation & Dice & Correlation \\
\hline
Inter-Rater & $0.687$ & n/a & n/a & n/a & n/a & n/a \\
NIH SSL & $0.512$ & $0.913$ & $0.690$ & $0.752$ & $0.601$ & $0.832$ \\
VUMC SSL & $0.407$  & $0.859$ & $0.745$ & $0.754$ & $0.576$ & $0.807$ \\
MSL & $0.552^*$ & $\textbf{0.943}$ & $0.725$ & $\textbf{0.791}$ & $0.63^*$ & $\textbf{0.867}$ \\
\bottomrule[2pt]
\end{tabular}
\end{center}
\end{table}

\section{Discussion}
To our best knowledge, this is the first application of multi-site distributed learning applied to clinical imaging data from different institutions.  In this paper, we have presented and validated a technique to distributively train a convolutional neural network over disparate data housed at different institutions.  While the multi-site model outperformed its single-site counterparts, our main contribution is a general framework to allow a neural model to train over more data than it would normally have access to while still preserving PHI.  We show that for this task, multi-site learning did not detract from the network's ability to learn over tasks, and as expected, performance improved with more data availability.  Additionally, our implementation to transfer the weights between sites automatically is straightforward, publicly available and can be generally applied to other epoch-based training scenarios.  Future work includes exploring alternate neural architectures such as U-net and evaluating the generalizablility of the MSL model compared with the SSL models using more than two sites.

\section{ACKNOWLEDGEMENTS}
Support for this work included funding from the Intramural Research Program of the NIH Clinical Center and the Department of Defense in the Center for Neuroscience and Regenerative Medicine, and NIH grants 1R01EB017230-01A1 (Landman) and 1R01GM120484-01A1 (Patel), as well as NSF 1452485 (Landman). The VUMC dataset was obtained from ImageVU, a research resource supported by the VICTR CTSA award (ULTR000445 from NCATS/NIH), Vanderbilt University Medical Center institutional funding and Patient-Centered Outcomes Research Institute (PCORI; contract CDRN-1306-04869). This work received support from the Advanced Computing Center for Research and Education (ACCRE) at the Vanderbilt University, Nashville, TN, as well as in part by ViSE/VICTR VR3029.  We also extend gratitude to NVIDIA for their support by means of the NVIDIA hardware grant.

%\bibliographystyle{spiebib}
%\small{
%\bibliography{remedios-spie2019-refs}
%}

\small{

}
\end{document}